\documentclass{article}

\usepackage{microtype}
\usepackage{graphicx}
\usepackage{subfigure}
\usepackage{booktabs} 

\usepackage{amsfonts} 
\usepackage{nicefrac} 
\usepackage{microtype} 
\usepackage{graphicx}
\usepackage{graphics}
\usepackage[table]{xcolor}
\usepackage{gensymb}
\usepackage{amsthm}
\usepackage{mathrsfs}
\usepackage{amsmath}
\usepackage{amssymb}
\usepackage{mathtools}
\usepackage{multirow}
\usepackage{soul}
\usepackage{multicol}
\makeatletter
\def\BState{\State\hskip-\ALG@thistlm}
\makeatother
\newcommand{\cdotv}{\boldsymbol{\cdot}}
\newtheorem{theorem}{Theorem}

\newtheorem{remark}[theorem]{Remark}

\usepackage{makecell}
\newcommand{\E}{\mathbb{E}}

\newcommand{{\phivmat}}{\boldsymbol{{\phi}}}
\newcommand{{\phiv}}{\boldsymbol{{\phi}}}
\newcommand{\ba}[1]{\begin{align}#1\end{align}}

\usepackage{stackengine}

\makeatletter
\newcommand{\distas}[1]{\mathbin{\overset{#1}{\kern\z@\sim}}}%

\newcommand{\beqs}{\vspace{0mm}\begin{eqnarray}}
\newcommand{\eeqs}{\vspace{0mm}\end{eqnarray}}
\newcommand{\barr}{\begin{array}}
\newcommand{\earr}{\end{array}}

\newcommand{\hv}[0]{{\boldsymbol{h}} }

\newcommand{\kv}[0]{{\boldsymbol{k}} }

\newcommand{\xv}{\boldsymbol{x}}
\newcommand{\yv}{\boldsymbol{y}}

\newcommand{\alphav}{\boldsymbol{\alpha}}
\newcommand{\betav}[0]{{\boldsymbol{\beta}}}

\newcommand{\epsilonv}{\boldsymbol{\epsilon}}

\newcommand{\etav}[0]{{\boldsymbol{\eta}}}

\newcommand{\thetav}{\boldsymbol{\theta}}

\newcommand{\lambdav}[0]{{\boldsymbol{\lambda}}}

\usepackage[textwidth=0.8in,textsize=tiny]{todonotes}

\newcommand{\given}{\,|\,}

\usepackage{hyperref}

\usepackage[accepted]{icml2021}

\newcommand{\ours}{{BABN}}

\begin{document}

\twocolumn[
\icmltitle{Bayesian Attention Belief Networks}




\icmlsetsymbol{equal}{*}

\begin{icmlauthorlist}
\icmlauthor{Shujian Zhang}{equal,to}
\icmlauthor{Xinjie Fan}{equal,to}
\icmlauthor{Bo Chen}{goo}
\icmlauthor{Mingyuan Zhou}{to}
\end{icmlauthorlist}

\icmlaffiliation{to}{The University of Texas at Austin}
\icmlaffiliation{goo}{Xidian University}

\icmlcorrespondingauthor{Mingyuan Zhou}{mingyuan.zhou@mccombs.utexas.edu}

\icmlkeywords{Machine Learning, ICML}

\vskip 0.3in
]



\printAffiliationsAndNotice{\icmlEqualContribution} 

\begin{abstract}
Attention-based neural networks have achieved state-of-the-art results on a wide range of tasks. Most such models use deterministic attention while stochastic attention is less explored due to the optimization difficulties or complicated model design. This paper introduces Bayesian attention belief networks, which construct a decoder network by modeling unnormalized attention weights with a hierarchy of gamma distributions, and an encoder network by stacking Weibull distributions with a deterministic-upward-stochastic-downward structure to approximate the posterior. The resulting auto-encoding networks can be optimized in a differentiable way with a variational lower bound. It is simple to convert any models with deterministic attention, including pretrained ones, to the proposed Bayesian attention belief networks. On a variety of language understanding tasks, we show that our method outperforms deterministic attention and state-of-the-art stochastic attention in accuracy, uncertainty estimation, generalization across domains, and robustness to adversarial attacks. We further demonstrate the general applicability of our method on neural machine translation and visual question answering, showing great potential of incorporating our method into various attention-related tasks.



\end{abstract}
\section{Introduction}
Attention-based architectures were originally proposed to induce useful inductive biases by aggregating features with learnable weights for sequence models  \cite{sutskever2014sequence,bahdanau2015neural}. Since the introduction of the attention-based Transformer \cite{vaswani2017attention}, attention has become the foundation for many state-of-the-art models. Due to the computational efficiency and scalability of the Transformer structure, it becomes possible to train unprecedented large models on big datasets \citep{devlin2018bert}, which stimulates a great amount of research to pretrain models on large unlabeled datasets. In an unsupervised manner, this approach learns useful representations that benefit downstream tasks, 
achieving tremendous success in natural language processing \citep{devlin2018bert,lan2019albert, liu2019roberta,joshi2020spanbert,radford2018improving,yang2019xlnet}, compute vision \citep{dosovitskiy2020image,chen2020generative}, and multi-modal tasks \citep{chen2019uniter,lu2019vilbert}. 

Most of the attention networks treat attention weights as deterministic rather than random variables, leading to the whole networks mostly composed of deterministic mappings. Such networks, although simple to optimize, are often incapable of modeling complex dependencies in data \cite{chung2015recurrent}.  By contrast, stochastic belief networks \citep{neal1992connectionist,hinton2006fast,gan2015learning,zhou2016augmentable,WHAI,fraccaro2016sequential,fan2021contextual, bayer2014learning,bowman2016generating}, stacking stochastic neural network layers, have shown great advantages over deterministic networks in not only modeling highly structured data but also providing uncertainty estimation. 

This paper proposes Bayesian attention belief networks (BABN), where we build deep stochastic networks by modeling unnormalized attention weights as random variables. First, we construct the generative (decoder) network with a hierarchy of gamma distributions. Second, the inference (encoder) network is a stack of Weibull distributions with a deterministic-upward and a stochastic-downward path. Third, we leverage the efficient structure of existing deterministic attention networks and use the keys and queries of current attention networks to parameterize the distributions of BABN. This efficient architecture design enables us to easily convert any existing deterministic attention networks, including pretrained ones, to BABN. Meanwhile, it imposes natural parameter and computational sharing within the networks, maintaining computation efficiency and preventing overfitting. Finally, we optimize both the decoder and encoder networks with an evidence lower bound. As the encoder network is composed of a reparameterizable distribution, $i.e.$, Weibull distribution, the training objective is differentiable. Further, leveraging the fact that the Kullback--Leibler (KL) divergence from the gamma to Weibull distribution is analytic, we can efficiently reduce the gradient estimation variance. 


The proposed BABN has a generic architecture so that any existing deterministic attention models, including pretrained ones, can be converted to BABN while maintaining the inherent advantages of conventional attention, such as efficiency and being simple to optimize. 
Our proposed method is generally simple to implement and boosts the performance while only slightly increasing the memory and computational cost.
On various natural language understanding tasks, neural machine translation, and visual question answering, our method outperforms vanilla deterministic attention and state-of-the-art stochastic attentions, in terms of accuracy and uncertainty estimation. We further demonstrate that {\ours} achieves strong performance in domain generalization and adversarial robustness.

\section{Background on Attention Networks}
Most attention structures can be unified with the key, query and value framework, where keys and queries are used to calculate attention weights and values are aggregated by the weights to obtain the final output. Formally, given $n$ key-value pairs and $m$ queries, we denote keys, values, and queries by $K\in \mathbb{R}^{n\times d_k}$,  $V\in \mathbb{R}^{n\times d_v}$, and $Q \in \mathbb{R}^{m\times d_k}$. Note that the second dimension of $K$ and $Q$ are often equal because we usually need to compute scaled dot-product between key and query \citep{vaswani2017attention} as $$\Phi = f_\text{dot}(Q,K) = QK^T/\sqrt{d_k}\in \mathbb{R}^{m\times n}.$$ To ensure that the attention weights are positive and sum up to one across keys, $f_\text{dot}$ is often followed by a softmax function to obtain the final attention weights $W=\text{softmax}(f_\text{dot}(Q,K))$. In detail, first we obtain positive unnormalized weights $S$ with the exponential function: $S = \exp(\Phi)$, then we normalize $S$ across the key dimension with $f_\text{norm}$ as $$W_{i,j} = f_\text{norm}(S)_{i,j} := \frac{S_{i,j}}{\sum_{j'=1}^n S_{i,j'}},$$ for $i=1,..., m, j=1,...,n.$ Finally, the output of attention is $O=WV\in \mathbb{R}^{m\times d_v}$,  aggregating the values according to the attention weights. 


This generic architecture can be used in many different models and applications. More interestingly, attention layers can be stacked {on top of each other} to build a deep neural network that is capable of modeling complicated deterministic functions. For example, in self-attention, denote the input of the $l$th attention layer by $I^l$, then we can obtain the key $K^l$, query $Q^l$, and value $V^l$ by linearly projecting $I^l$ to different spaces: $K^l= I^l M_{K}^l, Q^l= I^l M_{Q}^l, V^l= I^l M_{V}^l$, where $M$'s are parametric matrices to learn. The output of this attention layer, $O^l$, can be fed as next layer's input $I^{l+1}=O^l$, and we can iterate the above process to obtain a deep self-attention-based neural network. Note that other structure details \cite{vaswani2017attention}, such as residual structure \cite{he2016deep}, feed forward networks, and layer normalization \cite{ba2016layer}, are also indispensable for the network but it would not affect the general framework we describe here.

\section{BABN: Bayesian Attention Belief  Networks}

We introduce an efficient solution for deep attention belief networks: (a) build a hierarchical distribution to model unnormalized attention weights as the generative model, (b) develop an inference network with a deterministic-upward-stochastic-downward structure, and (c) leverage existing attention architectures and a few light-weight linear layers to parameterize the distributions. The resulting architecture can be efficiently learned with variational inference. 


\subsection{Deep Gamma Decoder Attention Networks}
Denoting a supervised learning problem with training data $\mathcal{D}:=\{\xv_i, \yv_i\}_{i=1}^N$, the conditional probability for conventional attention-based model is $p_{\thetav}(\yv_i\given \xv_i, W_i)$, where $W_i:=f_{\thetav}(\xv_i)$, $f_{\thetav}(\cdotv)$ is a deterministic transformation, and $\thetav$ is the neural network parameter that includes 
the attention projections $M$'s. For notational convenience, below we drop the data index $i$. 
Even though the deterministic attention mechanism is easy to implement and optimize, it often fails to capture complex dependencies or provide uncertainty estimation \cite{fan2020bayesian}.

To remedy such issues, we construct deep stochastic attention networks by treating attention weights as latent variables. Instead of directly modeling the normalized attention weights $W=\{W^l\}_{l=1}^L$ on the simplex, we find it easier to model the unnormalized weights $S=\{S^l\}_{l=1}^L$ on the positive real line.
We model the distribution of $S$ with a product of gamma distributions: 
$$p_{\etav}(S\given\xv)=\textstyle\prod_{l=1}^{L} \text{Gamma}(S^l\given \alphav^l=f_{\etav}^{l} (S^{1:l-1}, \xv), \betav),$$
where the shape parameter $\alphav^l$ at the $l$th layer is the output of a neural network $f_{\etav}^{l}$ parameterized by $\etav$, and the rate parameter is a positive constant $\betav$. The gamma distribution has been widely used for modeling positive real variables and is known to be capable of capturing sparsity and skewness. 
It is particularly attractive for modeling unnormalized attention weights because normalizing the gamma distributions with the same rate parameter leads to a  Dirichlet distribution, which is commonly used for modeling variables on the simplex \citep{blei2003latent,zhou2016augmentable,deng2018latent, fan2020bayesian}. 
In this way, the whole generative process can be expressed as:
$$S\sim p_{\etav}(\cdotv\given \xv),~~ \yv \sim p_{\thetav}(\cdotv \given\xv, f_\text{norm}(S)).$$

\begin{remark}
Bayesian inference via Gibbs sampling is available when $\{f_{\etav}^{l}\}_{l=1}^L$ are simple linear projections and $p_{\thetav}$ is the Poison distribution 
\cite{zhou2016augmentable}:
\begin{equation}
    \begin{split}
        & f_{\etav}^{l}(S^{1:l-1}, \xv)= W^l S^{l-1}, \text{ for } l=1,...,L, \\ 
        & \yv \sim \text{Poisson}(W^{L+1}S^L).
    \end{split}
\end{equation}
\end{remark}
We sketch the Gibbs sampler (see  \citet{zhou2016augmentable} for details) in Fig.~\ref{fig:sketch}, whose upward and downward structure
motivates the design of our encoder (inference) network architecture which we will discuss in detail in Section~\ref{sec:weibull}. 

{\bf Efficient and Expressive Structures for $\alphav^l$. }To be able to model complicated dependencies, we use neural networks to model the mapping $\{f_{\etav}^{l}\}_{l=1}^L$ from $S^{1:l-1}$ and $\xv$ to $S^l$. However, having separate neural networks for each $f_{\etav}^{l}$ would lead to memory and computation redundancy as it does not exploit the hierarchical relationships among $\{f_{\etav}^{l}\}_{l=1}^L$. Therefore, we leverage the current attention's efficient structure, and note that the key $K^l$ at layer $l$ is a function output of previous attention weights $S^{1:l-1}$ and input $\xv$. This motivates us to make use of the key $K^l$ at layer $l$ to construct $f_{\etav}^{l}$. In particular, we apply a two-layer MLP to transform key $K^l$ to obtain $\alphav^l$:
$$\alphav^l = \text{softmax}(f_{\etav,2}^l(\text{ReLU}(f_{\etav, 1}^l(K^l)))),$$ where $f_{\etav, 1}^l, f_{\etav, 2}^l$ are two linear layers connected by the nonlinear activation function, ReLU \citep{nair2010rectified}. This architecture imposes natural parameter and computation sharing in a hierarchical way, which could not only improve efficiency but also prevent overfitting. 
\subsection{Deep Weibull Encoder Attention Networks}\label{sec:weibull}
\begin{figure}[t] 
\centering
\includegraphics[width=1\columnwidth]{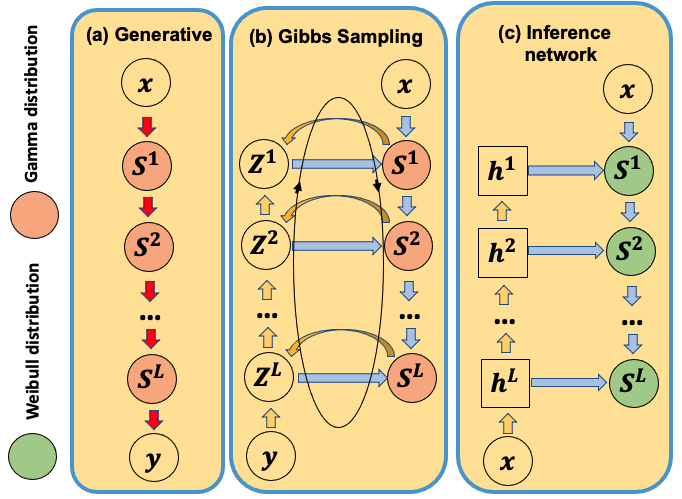}
\caption{(a) The structure of the generative model that models unnormalized attention weights with a hierarchy of gamma distributions. (b) A sketch of an upward-downward Gibbs sampler mimicking that of the gamma belief network \citep{zhou2016augmentable}, whose generative model is similarly structured  as in (a). 
$Z$ are augmented latent counts that facilitate the derivation of close-form Gibbs sampling update equations.  (c) Motivated by the Gibbs sampler's structure, we design the inference network in a similar upward-downward way, where $\hv$ represents a deterministic upward path and $S$ represents a stochastic downward path. Note that our inference network is not conditioned on $\yv$ as we are dealing with a supervised problem. Conditioned on $\yv$ would prevent directly using the inference network for new data points.} 
\label{fig:sketch} 
\end{figure}

\begin{figure*}[t] 
\centering
\includegraphics[width=16cm]{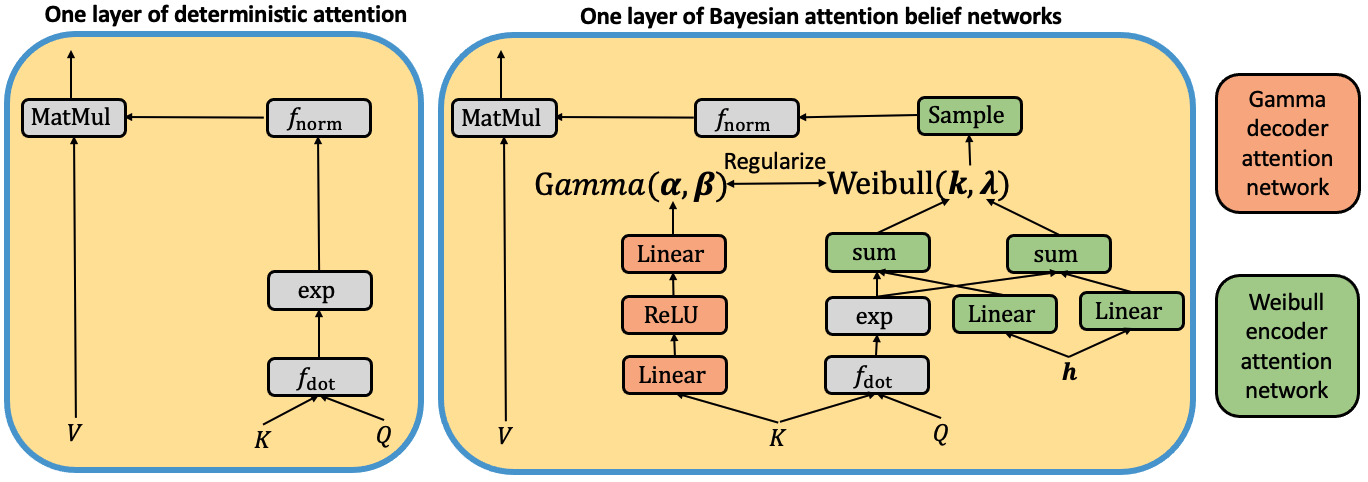} 
\caption{Illustration of the difference and similarity between the vanilla deterministic attention and one layer of our Bayesian attention belief networks. Bayesian attention belief networks (BABN) share the same architecture as the deterministic attention before obtaining key, query, and value. Then BABN adds light-weight linear layers to construct the gamma and Weibull distributions to model unnormalized attention weights, which are used after normalization to obtain the layer output as in the vanilla deterministic attention.} 
\label{fig:trans} 
\end{figure*}

Due to the nonlinear structure of the decoder attention network, deriving the Gibbs sampler is not feasible and its scalability is also a concern. In this regard, we propose an encoder network to learn a variational distribution $q_{\phiv}$ to approximate the posterior distribution of unnormalized attention weights $S$.

We model the variational distribution $q_{\phiv}$ with a product of Weibull distributions:
$$q_{\phiv}(S\given \xv, \yv) = \textstyle\prod_{l=1}^{L} \text{Weibull}(S^l\given \kv^l, \lambdav^l),$$
where $\kv^l, \lambdav^l$ are the Weibull shape and scale parameters, respectively.
The reason for choosing the Weibull distribution is threefold \citep{WHAI}: First, the Weibull is similar to gamma distribution, capable of modeling sparse, skewed, and positive distributions. Second, unlike the gamma distribution, the Weibull distribution has a simple reparameterization so that it is easier to optimize. That is, to sample $s\sim\text{Weibull}(k,\lambda)$ with probability density function (PDF) $p(s\given k,\lambda) = \frac{k}{\lambda ^k}s^{k-1}e^{-(s/\lambda)^k}$, it is equivalent to letting $S ={g}(\epsilon):= \lambda(-\log(1-\epsilon))^{1/k},~ \epsilon\sim \text{Unif}(0,1)$. Third, there exists an analytic KL divergence 
as
$
\small\textstyle\mbox{KL}(\text{Weibull}(k,\lambda)||\text{Gamma}(\alpha, \beta)) = \frac{\gamma \alpha}{k}-\alpha \log \lambda +\log k + \beta\lambda \Gamma(1+\frac 1k) - \gamma - 1-\alpha \log \beta+\log \Gamma (\alpha),
\nonumber$
where $\gamma$ denotes the Euler–Mascheroni constant and $\Gamma$ is the gamma function. This provides an efficient way to estimate the training objective which we will discuss in detail in Section~\ref{sec:vi}.




{\bf Deterministic-upward and Stochastic-downward Structure.} Inspired by the upward-downward Gibbs sampler sketched in Fig.~\ref{fig:sketch}, we mimic the structure to construct an inference network as: 
$$
\begin{array}{l}
\kv^l=f_{\kv, \hv}^{l}(\hv^l)+ f_{\kv, S}^{l}(S^{1:l-1},\xv),\\
\lambdav^l= f_{\lambdav, \hv}^{l}(\hv^l)+ f_{\lambdav, S}^{ l}(S^{1:l-1},\xv),\\
\hv^l= f_{\hv}^{l}(\hv^{l+1}),
\end{array}
$$
where $\{\hv^l\}_{l=1}^{L+1}$ serve as the augmented latent variables passing the information from data upwards and complement the downward information from attention variables $S$. 
A similar bottom-up and top-down structure was proposed in the ladder VAE \citep{sonderby2016ladder} and was found to help the optimization. In our experiments (section~\ref{sec:vqa}), we also found that the upward and downward structure plays an important role as the downward path delivers the prior information and the upward path delivers the likelihood information. Without the upward path of $\hv$, the model often has unstable performances. We note that although $q_{\phiv}$ is independent of $\yv$ during testing, it is possible for $q_{\phiv}$ to depend on part of $\yv$ that has already been observed by the model during training in sequence generation tasks, such as neural machine translation, where the queries come from $\yv$. Further, we think it is possible for $q_{\phiv}$ to approximate $p(S|\xv,\yv)$ even without conditioning on $\yv$ as $\xv$ conveys information of $\yv$. Formally, we define $f_{\kv, \hv}^{l}, f_{\kv, S}^{l}, f_{\lambdav, \hv}^{l}, f_{\lambdav, S}^{l}, f_{\hv}^{l}$ as follows: 
$$
\begin{array}{l}
\kv^l=\rho * \ln \left[1+\exp \left(f_{\phiv,1}^{l} (\hv^l)\right)\right] + \exp (\Phi^l),\\
\lambdav^l= \sigma *\ln \left[1+\exp \left(f_{\phiv,2}^{l} (\hv^l)\right)\right] + \frac{\exp (\Phi^l)}{  \Gamma(1 + 1/\kv^{l})},\\
\hv^l=\ln \left[1+\exp \left(f_{\phiv,3}^{l} (\hv^{l+1})\right)\right], 
\end{array}
$$
$\text{where }f_{\phiv,1}^{l}, f_{\phiv,2}^{l},\text{and } f_{\phiv,3}^{l}$ are linear layers that preserve the dimension of $\hv^l$, and $\hv^{L+1}$ is initialized as a function of $\xv$: $\hv^{L+1}=f_{\phiv, 0}(\boldsymbol{x})$. The structure involves the following parts.  
1) For $\kv^l,\lambdav^l$, we introduce weights $\rho, \sigma$ to balance the importance of the two parts in $\kv^l, \lambdav^l$.
2) We leverage the efficient deterministic attention architecture to construct the functions $f_{\kv, S}^{l}$ and $f_{\lambdav, S}^{l}$, where $\Phi^l=f(Q^l,K^l)$ is the function of $S^{1:l-1}$ and $\xv$. Using $\Phi^l$ to construct the inference network is an efficient way to introduce parameter and computation sharing between the layers of the encoder and decoder. 
3) For $\lambdav^l$, we rescale $\exp(\Phi^l)$ with $\Gamma(1+1/\kv^l)$ so that the expectation of the Weibull distribution is $\exp(\Phi^l)$ when $\sigma=0$, which corresponds to the deterministic attention before normalization. 
4) In addition, we model the functions $f_{\kv, \hv}^{l},f_{\lambdav, \hv}^{l},f_{\hv}^{l}$ with linear layers coupled with $\ln[1+\exp(\cdotv)]$ to obtain positive outputs. We need to point out that both $\Phi^l$ and $\hv^l$ are functions of only $\xv$ but not $\yv$, which enables us to directly use the variational distribution  $q_{\phiv}$ during testing for new data points \cite{wang2020thompson,fan2021contextual}. 5) We leverage the key and query of the {\it first} attention layer to initialize hidden states $\hv^{L+1}$. In particular, we let $\hv^{L+1}=\text{softmax}(\Phi^1)$. As there is yet no randomness introduced to $\Phi^1$, this mapping from $\xv$ to $\hv^{L+1}$ is still deterministic. 
By sharing the parameter and computation with the main network, $f_{\phiv, 0}$ does not add any memory or computation cost.

\begin{remark}\label{rem:1}
As our model leverages the efficient structure of the existing deterministic attention module and uses keys and queries to construct the prior and variational distribution for unnormalized attention weights, it is simple to convert existing deterministic attention networks to {\ours}. Fig.~\ref{fig:trans} shows that {\ours} shares parts of architecture with the deterministic attention. BABN adds a few light-weight linear layers to construct the gamma prior and Weibull variational distribution with the upward-downward structure.
More importantly, we note that we can use pretrained deterministic attention model checkpoints to initialize {\ours}, and then finetune the stochastic neural network. 
\end{remark}

\begin{remark}
{\ours} can be easily extended to multi-head attention, where queries, keys, and values are projected $H$ times linearly with $H$ different learned projections, and the outputs of $H$ heads are concatenated as the final output. Since the unnormalized multi-head attention weights are conditionally independent, we can still model the unnormalized attention weights with the same hierarchical formulation. Specifically, for each layer, conditioned on previous layers, we obtain the queries, keys for multiple heads to construct the distributions for unnormalized attention weights of each head separately. Then, we normalize the attention weights for each head so that within each head, the attention weights sum to one across keys, which is the same as the vanilla multi-head attention model.
\end{remark}

\subsection{Learning Bayesian Attention Belief Networks}\label{sec:vi}
Now, we have defined the gamma decoder network and Weibull encoder network. We learn the encoder network $q_{\phiv}$ to approximate the posterior distribution $p(S\given \xv, \yv)$ by minimizing the KL divergence, $\mathcal{L}_\text{KL}=\text{KL}(q_{\phiv}(S)||p(S\given \xv, \yv))$,
which is equivalent to maximizing, \ba{ \textstyle 
\mathcal{L}(\xv, \yv):=\E_{q_{\phiv}(S)}\left[ \log p_{\thetav}(\yv\given \xv, S)\right]-\mbox{KL}(q_{\phiv}(S)||p_\etav(S)),\label{eq:elbo}\nonumber
}
an evidence lower bound (ELBO) \citep{hoffman2013stochastic,blei2017variational, kingma2013auto} of the intractable log marginal likelihood $\log p(\yv \given \xv)=\log \int p_{\thetav}(\yv\given \xv, S)p_{\etav}(S)d S$. The objective $\mathcal{L}$ consists of two parts: the likelihood part, which maximizes the data likelihood under the encoder network; the regularization part, which enforces the variational distribution to be close to the prior distribution.
We also use the same objective $\mathcal{L}$ to learn the decoder networks $p_{\etav}$ and $p_{\thetav}$, as the exact marginal likelihood is intractable, and the ELBO is a good approximation when the variational distribution well approximates the true posterior \cite{kingma2013auto}.

Note that as $q_{\phiv}$ is a product of Weibull distributions, it is reparameterizable. In particular, to sample $S$ from $q_{\phiv}$, we sequentially sample $S^l$ conditional on previous samples $S^{1:l-1}$, as
$S^l\sim\text{Weibull}(S^l\given \kv^l, \lambdav^l)$. This can be realized by letting $S^l= g_{\phiv}^{l}(\epsilonv^l):= \lambdav^{l}(-\log (1-\epsilonv^l))^{1/{\kv^l}}$, where $\epsilonv^l$ is a tensor with the same shape as $S^l$ and its elements are $i.i.d$ samples from the uniform distribution. In practice, we found that drawing $\epsilonv^l$ from Uniform $(0,1)$ leads to numerical issues. Therefore, to prevent numeral instability, we choose to draw $\epsilonv^l$ from Uniform $(0.1, 0.9)$ as an approximation. Further, we note that at each layer $l$, the KL between the conditional distribution of encoder and decoder, $\text{KL}(q_{\phiv}(S^l\given S^{1:l-1}) || p_{\etav}(S^l\given S^{1:l-1}))$, is analytical. Therefore, we follow the same way in \citet{fan2020bayesian} to 
efficiently compute 
$\mbox{KL}(q_{\phiv}(S)||p_\etav(S)$ by decomposing it as 
$ 
        \sum\nolimits_{l=1}^L \E_{q_{\phiv}(S^{1:l-1})}\small\underbrace{{\text{KL}}(q_{\phiv}(S^l|S^{1:l-1})||p_\etav(S^l|S^{1:l-1}))}_{\text{analytic}},\nonumber
$ 
where the integrand is analytic. Putting it all together, we can rewrite the ELBO objective as
$\mathcal{L}(\xv,\yv)=\E_{\epsilonv}[\mathcal{L}_{\epsilonv}(\xv,\yv,\epsilonv)]$, where
\begin{equation}
  \begin{split}
    &\mathcal{L}_{\epsilonv}(\xv,\yv,\epsilonv)=
\log p_{\thetav}(\yv\given \xv, {g}_{\phiv}(\epsilonv)) \\
    &- \sum\nolimits_{l=1}^L\small \underbrace{\mbox{KL}(q_{\phiv}(S^l\given {g}_{\phiv}(\epsilonv^{1:l-1}))||p_\etav(S^l\given {g}_{\phiv}(\epsilonv^{1:l-1})))}_{\textit{analytic}}. \nonumber
  \end{split}
\end{equation}
With the reparameterization, now we can efficiently estimate the gradient of $\mathcal{L}$ with respect to $\thetav, \phiv, \etav$ by computing the gradient of $\mathcal{L}_{\epsilonv}$ with one sample of $\epsilonv$. Both reparameterization and semi-analytic KL \cite{mcbook} reduce the Monte Carlo estimation variance and still keep the estimation unbiased. Finally, following previous work 
\cite{bowman2016generating}, we add a weight $\lambda$
to the KL term and anneal it
from a small value to one.

\section{Related Work}
{\it Stochastic attentions: } \citet{xu2015show}, along with several following work \citep{shankar2018posterior,deng2018latent}, proposed hard attention to model attention weights with categorical distributions, which only attends to one subject at a time. The categorical distribution, however, is not reparameterizable and therefore hinders the use of standard backpropagation. REINFORCE gradient estimator makes the optimization possible, but it has high variance and one often needs to carefully design baselines to make the performance comparable to deterministic attention \citep{xu2015show,deng2018latent}. Stochastic soft attention, on the other hand, is less investigated. \citet{deng2018latent} proposed modeling attention weights with the Dirichlet distribution, which is not reparameterizable and introduces optimization difficulties. \citet{fan2020bayesian} considered using reparameterizable distributions, such as Lognormal and Weibull distributions, to model unnormalized attention weights, which alleviates the optimization issue of previous stochastic attention. Compared to \citet{fan2020bayesian} who try to convert deterministic attention modules to stochastic ones, our method is motivated from building a deep stochastic network by modeling attention weights as random variables. With a deterministic-upward and stochastic-downward structure, our inference network comprises Weibull distributions, whose scale parameter $\lambdav$ and shape parameter $\kv$ are both sample-dependent. This makes it differ from \citet{fan2020bayesian}, where the shape parameter $\kv$, controlling the uncertainty of distribution, is a hyperparameter and the inference network does not involve a deterministic upward path. The proposed generalization gives us greater flexibility in modeling attention weights. We also conduct more extensive experiments to investigate the domain generalization ability and adversarial robustness of stochastic attentions. 

{\it Deep stochastic networks: }Augmenting deterministic neural networks with random variables provides us a principled way to capture the randomness in data and estimate uncertainty \cite{gal2016dropout,chung2015recurrent,bowman2016generating,tran2018bayesian}. More importantly, stacking stochastic layers into a deep stochastic network instead of a shallow probabilistic model 
is often preferable due to its capability to model more complicated dependencies \cite{WHAI}. For example, \citet{WHAI} have applied a gamma belief network for topic modeling, and a deep Weibull  network is used to approximate the posterior for scalable inference. We apply a similar structure to the widely used attention models and leverage the existing efficient attention architecture to build scalable networks.

\begin{table*}[htp!]
\caption{Results of the in-domain accuracies for different models on GLUE and SQuAD benchmarks.}
\label{tab:nlp}
\begin{center}
\begin{sc}
\resizebox{2.05\columnwidth}{!}{
\begin{tabular}{@{}ccccccccccccc@{}}\toprule
Model & MRPC & CoLA & RTE & MNLI & QNLI  & QQP & SST-2 & STS & SQuAD 1.1 & SQuAD 2.0 \\ \midrule
ALBERT-base & 86.5 & 54.5 & 75.8 & 85.1 & 90.9 &  90.8 & 92.4 & 90.3& 80.86/88.70& 78.80/82.07\\
ALBERT-base+BAM & 88.5 & 55.8 & 76.2 & 85.6 & 91.5 & { 90.7} & 92.7 & 91.1 &  81.40/88.82 & 78.97/82.23\\
ALBERT-base+{\bf {\ours}} & {\bf 89.2}\footnotesize{$\pm$0.3}& {\bf 56.8}\footnotesize{$\pm$0.5 }& {\bf 77.6}\footnotesize{$\pm$0.6 }& {\bf 86.2}\footnotesize{$\pm$0.3} &{\bf 91.9}\footnotesize{$\pm$0.3 }& {\bf 91.2} \footnotesize{$\pm$0.1} & {\bf 93.1}\footnotesize{$\pm$0.2} &{\bf 91.8}\footnotesize{$\pm$0.2} & {\bf 81.81}\footnotesize{$\pm$0.1}/\normalsize {\bf89.10}\footnotesize$\pm$0.1 & {\bf 79.20}\footnotesize{$\pm$0.1} / \normalsize {\bf82.41}\footnotesize{$\pm$0.1}\\
\bottomrule
\end{tabular}}
\end{sc}
\end{center}
\end{table*}

\section{Experimental Results}


Our method can be straightforwardly deployed wherever the regular attention is utilized. To test its effectiveness and general applicability, we apply our method to a diverse set of tasks, including language understanding, neural machine translation, and visual question answering.
For language understanding, we further study a model's generalization across domains and robustness towards adversarial attacks. Meanwhile, we experiment with a diverse set of state-of-the-art models, including, ALBERT \cite{lan2019albert}, BERT \cite{devlin2018bert}, and RoBERTa \cite{liu2019roberta}.  In the following, we provide the main experimental settings and results, with more details provided in Appendix \ref{sec:app_exp}.



\subsection{Attention in Natural Language Understanding}\label{sec:bert}
The self-attention-based Transformer models have become the de-facto standard for NLP tasks. The dominant approach is to first pretrain models on big corpora to learn generic features and then finetune the models on the corresponding datasets for downstream tasks. This approach has constantly been refreshing the state-of-the-art results on various tasks. However, the cost of training such models from scratch is often prohibitive for researchers with limited resources and it also brings burdens to our environment \cite{strubell2019energy}. For example, it takes $79$ hours to train a BERT-base model on $64$ V100 GPUs, which costs about $\$3,751$-$\$12,571$ cloud computations and brings CO$_2$ emissions of $1438$ lbs \cite{strubell2019energy}. Considering this, we believe that starting from pretrained models is not only efficient and environmental friendly, but also makes it accessible for researchers with limited computations. As discussed in Remark~\ref{rem:1}, we can convert a pretrained deterministic attention model to {\ours} and then finetune it on downstream tasks. Therefore, in this section, we investigate the effectiveness of only applying {\ours} during the finetuning stage.

\subsubsection{In-domain Performance Evaluation}\label{sec:indomain}
First, we consider the standard setting, $i.e.$, evaluating in-domain accuracies, where both the training and testing data are from the same domain.

{\bf Experimental Settings.} 
We include $8$ datasets from General Language Understanding Evaluation (GLUE) \cite{wang2018glue} and two versions of Stanford Question Answering Datasets (SQuAD) \cite{rajpurkar2016squad,rajpurkar2018know} as the benchmarks. 
We build our method on a state-of-the-art model, ALBERT \cite{lan2019albert}, which is a memory-efficient version of BERT \cite{devlin2018bert} with parameter sharing and embedding factorization. We leverage the pretrained checkpoint as well as the codebase for finetuing provided by Huggingface PyTorch Transformer \cite{wolf2019transformers}. We use the base version of ALBERT \cite{lan2019albert}. During testing, we obtain point estimates by approximating the posterior means of prediction probabilities by substituting the latent unnormalized attention weights by their posterior expectations \cite{srivastava2014dropout}. 

\textbf{Results. }In Table~\ref{tab:nlp}, we compare BABN with the deterministic attention and BAM \cite{fan2020bayesian}, which is the state-of-the-art stochastic attention. BAM is also applied during the finetuning stage, resuming from the same checkpoint. We report the mean accuracies and standard deviations for $5$ independent runs. Table~\ref{tab:nlp} shows that BABN outperforms both deterministic attention and  BAM, which indicates that stochastic belief networks give better performance than deterministic ones and the more flexible structure of {\ours} is also preferable to the structure of BAM.
We consistently observe clear 
improvements 
even though we only apply {\ours} at the finetuning stage.\footnote{We provide the parameter sizes and step time for different attention types combined with ALBERT-base, a Transformer-based model, where the attention module constructs the main model in Table \ref{tab:glue_parameter} in the Appendix.} We leave as future work using {\ours} at the pretrain stage. 


\subsubsection{Generalization across Domains}\label{sec:outdomain}
In real applications, it is very likely to apply a deep learning model to the data from a new domain unseen in the training dataset. Therefore, it is important to evaluate a model's generalization ability across domains. In NLP, significant work has studied domain generalization on sentiment analysis \citep{chen2018adversarial, peng2018cross, miller2019simplified}. Recently,  \citet{desai2020calibration} studied the cross-domain generalization of pretrained Transformer models on more difficult tasks and found it still challenging for these pretrained models to generalize. In this section, we follow the setting of \citet{desai2020calibration} to study the generalization ability of our method. 

{\bf Experimental Settings.} Following \citet{desai2020calibration}, we test domain generalization on three challenging tasks, including natural language inference (NLI), paraphrase detection (PD), and commonsense reasoning (CR). Each task includes both a source domain, used for finetuning the model, and a target domain, used for evaluating the model. Specifically, SNLI \citep{bowman2015large} and MNLI \citep{williams2018broad} are the source and target domains for NLI, respectively; QQP and TwitterPPDB \citep{lan2017continuously} are the source and target domains for PD,  respectively; SWAG \citep{zellers2018swag} and HSWAG \citep{zellers2019hellaswag} are the source and target domains for CR,  respectively. These benchmarks are known to exhibit challenging domain shifts \citep{desai2020calibration}.  
For each experiment, we report both the in-domain (ID) accuracy on the source domain and out-of-domain (OD) accuracy on the target domain. As in \citet{desai2020calibration}, we also report the expected calibration error (ECE) as a measure of model calibration. 
To compute ECE, we need to divide the samples into groups with their confidences, defined as the probability of the maximum predicted class. Then, ECE$:= \sum_i \frac{B_i}{N} |\text{acc}(B_i) - \text{conf}(B_i)|$, where $B_i$, acc$(B_i)$, and conf$(B_i)$ are the count, accuracy, and confidence of samples in the $i$th group, respectively. We set the number of groups to $10$ as in \citet{desai2020calibration}.

\begin{table}[t!]
\caption{\small Results of domain generalization. We report the accuracy and ECE of various models on both in-domain data and out-of-domain data for three tasks: natural language inference, paraphrase detection, and commonsense reasoning.}\vspace{-3.5mm}
\label{tab:out_of_domain}
\begin{center}
\begin{sc}
\resizebox{1\columnwidth}{!}{
\begin{tabular}{@{}lccccc@{}}\Xhline{3\arrayrulewidth}
\small  & \multicolumn{2}{c}{\small Accuracy $\uparrow$} & 
& \multicolumn{2}{c}{\small ECE $\downarrow$}\\
\cmidrule{2-3} \cmidrule{5-6}
\small  & \small ID & \small OD && \small ID  & \small OD  \\ \Xhline{3\arrayrulewidth}
\small {Natural Language Inference} &SNLI&MNLI&&SNLI&MNLI \\ \midrule
\small DA \citep{parikh2016decomposable} & 84.63\small & 57.12 && {\bf1.02} & 8.79 \\
\small ESIM \citep{chen2017enhanced}& 88.32\small & 60.91 && 1.33 & 12.78 \\
\small BERT-base \citep{desai2020calibration} & 90.04\small & 73.52 && 2.54 & 7.03 \\
\small BERT-base+BAM & 90.25\small  & 73.81  &&2.37  & 6.40 \\
\small BERT-base+{\bf {\ours}} & {\bf90.63} & {\bf74.32} && 1.98 & \bf{5.09} \\ \midrule
\small RoBERTa-base  & 91.23\small & 78.79 && {\bf1.93} & 3.62 \\
\small RoBERTa-base+BAM  & 91.29 \small & 79.11  && 2.85 & 2.94 \\
\small RoBERTa-base+{\bf {\ours}} & {\bf91.70} & {\bf79.86} && 2.62 & {\bf2.67}\\\Xhline{3\arrayrulewidth}
\small {Paraphrase Detection}& QQP &Twitter & &QQP & Twitter \\ \midrule
\small DA \citep{parikh2016decomposable} & 85.85\small & 83.36 && 3.37 & 9.79 \\
\small ESIM \citep{chen2017enhanced}& 87.75\small & 84.00 && 3.65 & 8.38 \\
\small BERT-base \citep{desai2020calibration}& 90.27\small & 87.63 && 2.71 & 8.51 \\
\small BERT-base+BAM & 90.77\small &87.14   && 2.91 & 9.21 \\
\small BERT-base+{\bf {\ours}} & {\bf90.84} & {\bf 88.32} && {\bf1.42} & \bf{7.43} \\ \midrule
\small RoBERTa-base \citep{desai2020calibration}& 91.11\small & 86.72 && 2.33 & 9.55 \\
\small RoBERTa-base+BAM  & 91.24\small & 86.87  && 2.01  & 9.50 \\
\small RoBERTa-base+{\bf {\ours}} & {\bf91.72} & {\bf87.31} && {\bf1.74} & {\bf9.42}\\\Xhline{3\arrayrulewidth}
\small {Commonsense Reasoning} & SWAG & HSWAG&&SWAG&HSWAG \\ \midrule
\small DA \citep{parikh2016decomposable} & 46.80\small & 32.48 && 5.98 & 40.37 \\
\small ESIM \citep{chen2017enhanced}& 52.09\small & 32.08 && 7.01 & 19.57 \\
\small BERT-base \citep{desai2020calibration}& 79.40\small & 34.48 && 2.49 & 12.62 \\
\small BERT-base+BAM & 79.44\small  & 35.18  &&  2.38& 12.49 \\
\small BERT-base+{\bf {\ours}} & {\bf 79.57} & {\bf36.23} && {\bf1.91} & \bf{11.82} \\ \midrule
\small RoBERTa-base \citep{desai2020calibration}& 82.45\small & 41.68 && 1.76 & 11.93 \\
\small RoBERTa-base+BAM  & 82.61\small & 42.04  && 1.66  & 11.21 \\
\small RoBERTa-base+{\bf {\bf {\ours}}} & {\bf83.12} & {\bf43.11} && {\bf1.32} & {\bf9.72}\\
\Xhline{3\arrayrulewidth}
\end{tabular}}
\end{sc} 
\end{center}
\end{table}

\textbf{Results. }We summarize our results in Table~\ref{tab:out_of_domain}. Our baselines include two small-scale and non-pretrained models: Decomposable Attention (DA) \citep{parikh2016decomposable} and Enhanced Sequential Inference Model (ESIM) \citep{chen2017enhanced}, and two state-of-the-art large-scale and pretrained models with deterministic attention: BERT-base \cite{devlin2018bert} and RoBERTa-base models \cite{liu2019roberta}. We experiment with adding {\ours} to both BERT-base and RoBERTa-base models. Table~\ref{tab:out_of_domain} shows that adding BABN consistently improves upon the corresponding deterministic models on not only in-domain, which confirms our results in Section  \ref{sec:indomain}, but also out-of-domain. The performance gains on out-of-domain are often greater than the gains on in-domain, meaning that BABN can significantly help the model to generalize across domains. This gets along with our intuition that deep stochastic models should generalize better than deterministic ones. Further, we note that {\ours} also improves ECE, meaning that {\ours} helps to obtain better-calibrated models for uncertainty estimation.

\subsubsection{Robustness towards Adversarial Attacks}
Neural networks are known to be vulnerable to adversarial examples that have imperceptible perturbations from the original counterparts \citep{goodfellow2014explaining}. It has been found that even large language models pretrained on large corpora still suffer from the same issue \cite{jin2020bert}. Therefore, it is important to evaluate and improve a model's robustness against adversarial attacks. We argue that as our Bayesian attention belief networks are built by stacking probabilistic layers, the stochastic connections would make the model more robust so that it is more difficult to generate perturbations that would fool our model. 

{\bf Experimental Settings.} To compare the adversarial robustness of {\ours} and the deterministic attention, we first finetune the ALBERT-base models according to the same settings as in Section \ref{sec:indomain}, and then apply three state-of-the-art untargeted black-box adversarial attacks, including (1) Textfooler \citep{jin2020bert}, generating natural looking attacks with rule-based synonym replacement; (2) Textbugger \citep{li2019textbugger},  generating misspelled words by character- and word-level perturbations; (3) BAE \citep{garg2020bae}, generating BERT-based adversarial examples.  We implement all the attacks using the NLP attack package, TextAttack \citep{morris2020textattack}, with the default settings. For each model, we conduct $1000$ adversarial attacks and Table~\ref{tab:adv} reports the percentages of failed adversarial attacks. Higher percentages indicate more robust models.

\begin{table}[t!]
\caption{Results of pretrained large-scale models' robustness against adversarial attacks. For each model, we report the percentages of failed attacks under three adversarial attacks respectively. }
\label{tab:adv}
\begin{center}
\begin{sc}
\resizebox{1\columnwidth}{!}{
\begin{tabular}{@{}llcccccc@{}}\toprule
      {Attack} & {Attention} & {MRPC} & {CoLA} & {RTE} & {QQP} & {SST-2} & {Avg.}  \\ \midrule
\multirow{3}{*}{Textfooler} & base & {\bf 6.5} & 2.6 & 16.2 & 25.4 &7.0 & 11.5 \\ 
       & BAM & 6.2 & {3.1} & {\bf 17.8} & {28.7} & {12.5} & {12.5}\\ 
       & {\bf {\ours}} & 6.2 & {\bf 5.1} & {17.7} & {\bf 33.7} & {\bf 16.4} & {\bf 15.8}\\ 
      \midrule
      \multirow{3}{*}{Textbugger} & base & {\bf 10.6} & 16.8 & 19.9 & 30.1 & 40.1 & 23.5 \\ 
      & BAM & 9.9 & {16.7} & {21.0} & {32.5} & {51.7} & {26.4}\\
       & {\bf {\ours}} & 9.5 & {\bf 17.6} & {\bf 21.4} & {\bf 35.8} & {\bf 55.5} & {\bf 28.0}\\ 
      \midrule
      \multirow{3}{*}{BAE} & base & 44.8 & 4.9 & 35.6 & {\bf 48.8} & 13.9 & 29.6\\ 
      & BAM & 48.6 & {5.1} & {\bf 36.3} & {42.2} & {22.8} & {31.0}\\
       & {\bf {\ours}} & {\bf 50.4} & {\bf 7.1} & {35.9} & 42.8 & {\bf 25.7} & {\bf 32.4}\\ 
\bottomrule
\end{tabular}}
\end{sc}
\end{center}
\end{table}

{\bf Results.} Table~\ref{tab:adv} shows that {\ours} outperforms the deterministic attention baseline on most datasets, and achieves a much better average accuracy. The improvement is consistent across all three different adversarial attacks with different levels of failure rates, with Textfooler being the strongest attacker. These results verify our conjecture that by stacking stochastic layers, our Bayesian attention belief networks are more robust than deterministic models due to the stochastic connections. To the best of our knowledge, it is the first time to show that stochastic attention could improve adversarial robustness on large language models.

\subsection{Attention in Neural Machine Translation}
To show that {\ours} is generally applicable, we conduct experiments on the task of neural machine translation and compare BABN with SOTA stochastic attentions, including variational attention (VA) based methods \citep{deng2018latent} and BAM \citep{fan2020bayesian}. 

{\bf Experimental Settings.} For fair comparisons, we adapt the deterministic attention model used by \citet{deng2018latent} to BABN. The model is very different from the previous models, as it is LSTM-based, where attention is used to connect the encoder and decoder of the translation system \citep{deng2018latent}. We follow the experimental settings of \citet{deng2018latent}. Models are trained from scratch. IWSLT \cite{cettolo2014report} is used as benchmark. We adopt the widely used BLEU score \cite{papineni2002bleu} as the evaluation metric for the translation results. Experimental details are summarized in Appendix~\ref{sec:app_exp}.


\begin{table}[ht!]
\caption{\small Results of BLEU scores, parameter size and step time for different attentions on IWSLT.}\vspace{3mm}
\centering
\label{tab:nmt_accuracy}
\begin{sc}
\resizebox{1\columnwidth}{!}{\centering
\begin{tabular}{@{}lccccc@{}}\toprule
Attention & BLEU $\uparrow$ & Params $\downarrow$ &s/step $\downarrow$ \\ \midrule
Base  & 32.77 &42M & 0.08\\
VA  + enum \citep{deng2018latent} & 33.68 & 64M & 0.12\\
VA + sample \citep{deng2018latent}  & 33.30 & 64M & 0.15\\
BAM \citep{fan2020bayesian} & 33.81\small{$\pm0.02$}& 42M &0.10\\
{\bf {\ours}} & {\bf34.23}\small{$\pm0.05$} & 42M&0.11\\
\bottomrule
\end{tabular}}
\end{sc}
\end{table}

{\bf Results. }In Table~\ref{tab:nmt_accuracy}, we report the BLEU scores, model parameter sizes, and step time (second/step) for each attention type. It shows that {\ours} gives the best BLEU score outperforming deterministic attention (base), variational attention (VA), and BAM, while keeping the parameter size at the same level as deterministic attention. The runtime of {\ours} is on a par with BAM and slightly slower than deterministic attention, but it outruns the variational attention methods.

\subsection{Attention in Visual Question Answering}\label{sec:vqa}
We also conduct experiments on a multi-modal learning task, visual question answering (VQA) \cite{goyal2017making}, where the model learns to predict the answer to a given question on a given image. Transformer-like attention architectures have been widely  used to learn the multi-modal reasoning between image and language \cite{yu2019deep}. We adapt the recently proposed MCAN model \cite{yu2019deep} to BABN and compare with deterministic attention and BAM \cite{fan2020bayesian}.



{\bf Experimental Settings.} We mainly follow the setting by \citet{yu2019deep}, and experiment on the VQA-v2 dataset \cite{goyal2017making}. As in \citet{fan2020bayesian}, we also include a noisy dataset by perturbing the input with Gaussian noise to the image features \cite{larochelle2007empirical} to investigate the model’s robustness. We use $4$-layer encoder-decoder based MCAN as the baseline model, where the deterministic attention was originally used. We report accuracies as well as uncertainty estimations, which are measured by a hypothesis testing based Patch Accuracy vs Patch Uncertainty (PAvPU) \cite{fan2020bayesian,mukhoti2018evaluating}, reflecting whether the model is uncertain about its mistakes. The higher the PAvPU is, the better the uncertainty estimation is. We set the $p$-value threshold to be $0.05$ \citep{fan2020bayesian}. For uncertainty estimation, we sample $20$ unnormalized attention weights from the variational distribution. We provide more detailed experimental settings in Appendix~\ref{sec:app_exp}.


\begin{table}[t!]
\caption{\small Accuracies and PAvPUs of different attentions on both the original VQA-v2 dataset and the noise ones.}
\label{tab:vqa_accuracy}
\begin{center}
\begin{sc}
\resizebox{0.85\columnwidth}{!}{
\begin{tabular}{@{}lccccc@{}}\toprule
\small  & \multicolumn{2}{c}{\small Accuracy $\uparrow$} & 
& \multicolumn{2}{c}{\small PAvPU $\uparrow$}\\
\cmidrule{2-3} \cmidrule{5-6}
\small  & \small Original & \small Noisy && \small Original  & \small Noisy  \\ \midrule
\small Base & 66.74\small & 63.58 && 71.96 & 68.29 \\
\small BAM & 66.82\small & 63.98 && 72.01 & 68.58 \\
\small {\bf {\ours}} & {\bf66.92}\small{$\pm0.02$}& {\bf64.40}\small{$\pm0.03$} && {\bf72.21}\small{$\pm0.03$} & \bf{70.43}\small{$\pm0.04$}\\
\bottomrule
\end{tabular}}
\end{sc} 
\end{center}
\end{table}

\textbf{Results.} In Table~\ref{tab:vqa_accuracy}, we report the accuracy and PAvPU of different attentions on both original and noisy data. It shows that {\ours} consistently improve upon the deterministic attention and BAM in terms of both accuracy and PAvPU, meaning that {\ours} in general is more uncertain on its mistakes and more certain on its correct predictions. Further, we note that the performance gain is more significant on the noisy dataset, indicating that {\ours} helps to learn a more robust model, which also agrees with 
our results on domain generalization in Section   \ref{sec:outdomain}.

\begin{figure}[t] 
\centering
\includegraphics[width=8cm]{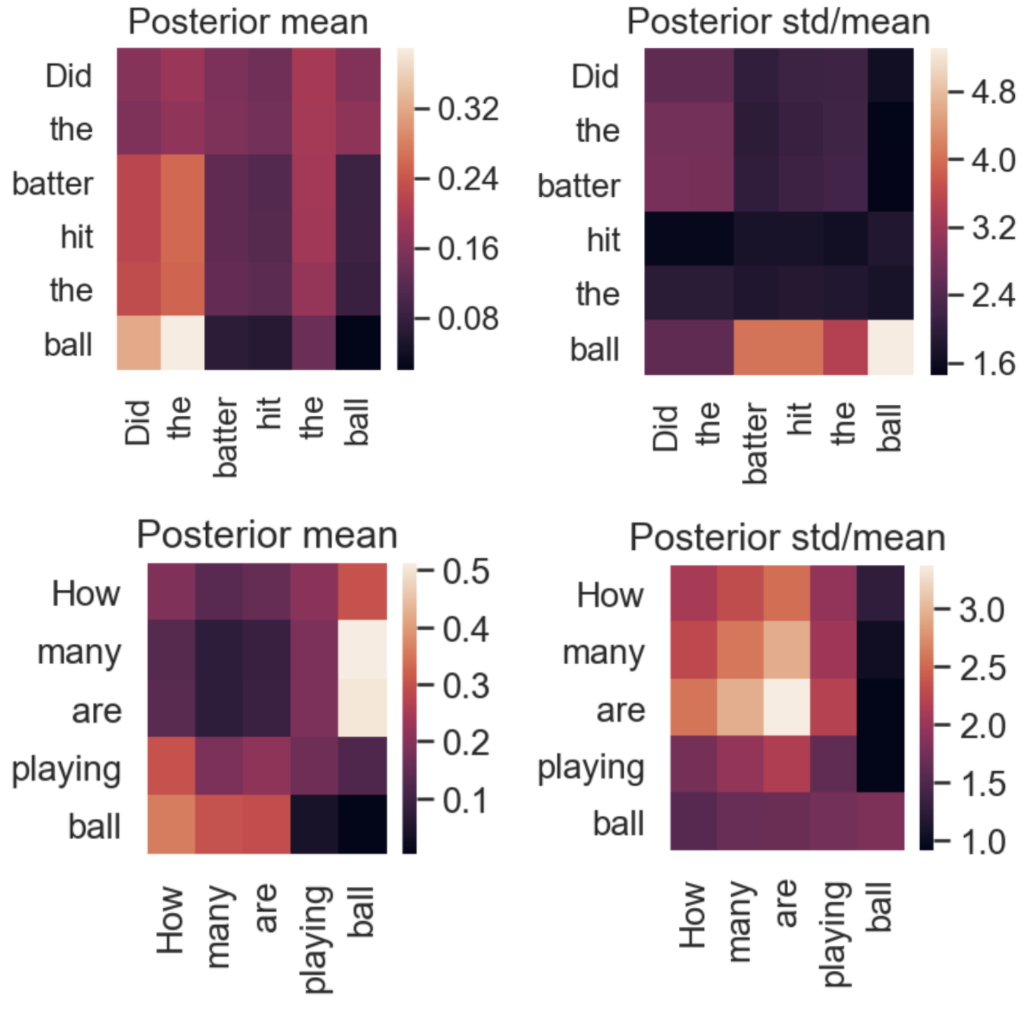}
\caption{\small For two questions from VQA, we visualize the posterior mean and std/mean for attention weights of {\ours},  where each row corresponds to one question. Rows represent queries, and columns represent keys. For example, considering the first question, on the left plot, when the row is ``Did'' and the column is ``hit'', the color represents the average attention weight from the query ``Did'' to the key ``hit''. On the right plot, the color at the same location represents the uncertainty from the query ``Did'' to the key ``hit''.
We note that the model is mostly certain except for the query ``ball'' from the right plot, which is assigning high average attention weights for ``Did'' and ``the'' rather than other words as shown on the left.  } 
\label{fig:att_vis1} 
\end{figure}

\textbf{Results Analysis. } \textit{Visualizations.} In Fig.~\ref{fig:att_vis1}, 
we plot statistics of the posterior distributions for the attention weights of one question in VQA. We visualize the normalized posterior mean (left) as a measure of the average importance of each query-key pair, and posterior standard deviation divided by posterior mean (std/mean on the right) as a measure of uncertainty. 
The plot shows that {\ours} is able to learn different uncertainties (std/mean) for each query-key pair in contrast to the fixed std/mean of BAM.
This sample-dependent uncertainty of {\ours} enables the strong capability in modeling attention weights and therefore gives good uncertainty estimation. 

\textit{Ablation Study. }We also conduct ablation study to exam the role of the upward-downward structure by turning the weight parameters $\rho$ and $\sigma$ to zeros. We found that tuning either parameter to zero would lead to performance drop, especially the parameter $\rho$, which demonstrates the necessity and effectiveness of the upward-downward structure. Please see detailed results in Table~\ref{tab:vqa_ablation} in Appendix.

\section{Conclusion}
We propose Bayesian attention belief network ({\ours}), a deep stochastic network by modeling attention weights as hierarchically dependent random variables. 
A multi-stochastic-layer generative model and a deterministic-upward-stochastic-downward inference network are constructed by leveraging the existing attention architecture. 
This generic and efficient architecture design enables us to easily convert existing deterministic attention models, including pretrained ones, to {\ours}, while only slightly increasing memory and computational cost. On various language understanding tasks, {\ours} exhibits strong performance in accuracy, uncertainty estimation, domain generalization, and adversarial robustness. Interestingly, clear improvement in performance has already been achieved by adding {\ours} only during the finetuning stage. 
We further demonstrate the general applicability of {\ours} on additional tasks, including neural machine translation and visual question answering, where BABN consistently outperforms corresponding baselines and shows great potential to be an efficient alternative to many existing attention models.


\section*{Acknowledgements}
S. Zhang, X. Fan, and M. Zhou acknowledge the support of Grants IIS-1812699 and ECCS-1952193
from the U.S. National Science Foundation, the APX 2019 project sponsored by the Office of the Vice
President for Research at The University of Texas at Austin, the support of a gift fund from ByteDance Inc., and the Texas Advanced Computing Center
(TACC) 
for providing HPC resources that have contributed to
the research results reported within this paper. 

\small
\bibliographystyle{icml2021}
\bibliography{reference.bib}
\normalsize

\clearpage
\appendix

\section{Experimental details}\label{sec:app_exp}

\subsection{Natural Language Understanding} 

\subsubsection{Model Specifications for In-domain Evaluation}

ALBERT \cite{lan2019albert} is used as the pretrained model on large corpora to extract the context embeddings. ALBERT is a memory-efficient version of BERT with parameter sharing and embedding factorization. In our experiments, we use the ALBERT-base model with $12$ attention layers and hidden dimension $768$. The embedding dimension for factorized embedding is $128$.

\subsubsection{Experimental Settings for In-domain Evaluation}

Our experiments are conducted on both the General Language Understanding Evaluation (GLUE) and Stanford Question Answering (SQuAD) Datasets. There are $8$ tasks in GLUE, including Microsoft Research Paraphrase Corpus (MRPC; \cite{dolan2005automatically}), Corpus of Linguistic Acceptability (CoLA; \cite{warstadt2019neural}), Recognizing Textual Entailment (RTE; \cite{dagan2005pascal}), Multi-Genre NLI (MNLI; \cite{williams2017broad}), Question NLI (QNLI; \cite{rajpurkar2016squad}), Quora Question Pairs (QQP; \cite{iyer2017first}), Stanford Sentiment Treebank (SST; \cite{socher2013recursive}), and Semantic Textual Similarity Benchmark (STS;\cite{cer2017semeval}). For SQuAD, we include both SQuAD v1.1 and SQuAD v2.0. We use the codebase\footnote{\url{https://github.com/huggingface/transformers}} from Huggingface Transformers \cite{wolf2019transformers}. For the detailed experimental settings, we summarize in  Table~\ref{tab:albert_setting}.

\begin{table}[htp!]\vspace{-5mm}
\caption{Experimental settings of each task for in-domain pretrained language model (LR: learning rate, BSZ: batch size, DR: dropout rate, TS: training steps, WS: warmping steps, MSL: maximum sentence length).}\vspace{-1mm}
\label{tab:albert_setting}
\begin{center}
\begin{small}
\begin{sc}
\resizebox{1.0\columnwidth}{!}{
\begin{tabular}{@{}ccccccccc@{}}\toprule
& \text { LR } & \text { BSZ } & \text { ALBERT DR } & \text { Classifier DR } & \text { TS } & \text { WS } & \text { MSL } \\ \midrule
\text { CoLA } & 1.00$e^{-5}$ & 16 & 0 & 0.1 & 5336 & 320 & 512 \\
\text { STS } & 2.00$e^{-5}$ & 16 & 0 & 0.1 & 3598 & 214 & 512 \\
\text { SST2 } & 1.00 $e^{-5}$ & 32 & 0 & 0.1 & 20935 & 1256 & 512 \\
\text { MNLI } & 3.00 $e^{-5}$ & 128 & 0 & 0.1 & 10000 & 1000 & 512 \\
\text { QNLI } & 1.00 $e^{-5}$ & 32 & 0 & 0.1 & 33112 & 1986 & 512 \\
\text { QQP } & 5.00 $e^{-5}$ & 128 & 0.1 & 0.1 & 14000 & 1000 & 512\\
\text { RTE } & 3.00 $e^{-5}$ & 32 & 0.1 & 0.1 & 800 & 200 & 512\\
\text { MRPC } & 2.00 $e^{-5}$ & 32 & 0 & 0.1 & 800 & 200 & 512\\
\text { SQuAD v1.1 } & 5.00 $e^{-5}$ & 48 & 0 & 0.1 & 3649 & 365 & 384 \\
\text { SQuAD } v2.0 & 3.00 $e^{-5}$ & 48 & 0 & 0.1 & 8144 & 814 & 512\\
\bottomrule
\end{tabular}}
\end{sc}
\end{small}
\end{center}
\end{table}

\begin{table}[htp]
\vspace{-10mm}
\caption{\scriptsize Efficiency on ALBERT-base models.}
\centering
\label{tab:glue_parameter}
\begin{sc}
\resizebox{0.45\columnwidth}{!}{\centering
\begin{tabular}{@{}lccccc@{}}\toprule
Attention & Params $\downarrow$ &s/step $\downarrow$ \\ \midrule
Base  & 11.7M & 0.26\\
BAM  & 11.7M & 0.35\\
{{\ours}} & 12.4M & 0.41\\
\bottomrule
\end{tabular}}
\end{sc}
\end{table}



\subsubsection{Model Specifications for Domain Generalizations}
We follow \citet{desai2020calibration} to
use bert-base-uncased \cite{devlin2018bert} and roberta-base \cite{liu2019roberta} as the baseline models. We also include the results of two non-pretrained models DA \cite{parikh2016decomposable} and ESIM \cite{chen2017enhanced} from \citet{desai2020calibration}, which are obtained with the open-source implementation in AllenNLP \cite{gardner2017deep}.
The pretrained models are provided by HuggingFace Transformers \cite{wolf2019transformers}. Largely following the settings from \citet{desai2020calibration}. we finetune BERT with a maximum of $3$ epochs, batch size of $16$, learning rate of 2$e^{-5}$,
gradient clip of $1.0$, and no weight decay. For RoBERTa, we finetune with a maximum of
$3$ epochs, batch size of $32$, learning rate of 1$e^{-5}$,
gradient clip of $1.0$, and weight decay of $0.1$. AdamW \cite{loshchilov2018decoupled} is used as the optimizer in experiments.  

\subsubsection{Experimental Settings for Domain Generalizations}
For all datasets, we follow the settings from \citet{desai2020calibration} and split the development
set in half to obtain a held-out, non-blind test set.

We conduct experiments on three tasks: (1) \textit{Natural Language Inference.} The Stanford Natural Language Inference (SNLI) corpus is a large-scale entailment dataset \cite{bowman2015large}. The similar entailment data across domains is also included in Multi-Genre Natural Language Inference (MNLI) \cite{williams2018broad}. Thus the MNLI can be used as an unseen out-of-domain test dataset.
(2) \textit{Paraphrase Detection.} Quora Question Pairs (QQP) contains sentence pairs from Quora that are
semantically equivalent \cite{iyer2017first}. TwitterPPDB (TPPDB), considered as out-of-domain data, contains the sentence pairs from the paraphrased tweets \cite{lan2017continuously}.
(3) \textit{Commonsense Reasoning.} Situations With Adversarial Generations (SWAG) is a grounded commonsense reasoning task \cite{zellers2018swag}. The out-of-domain data is HellaSWAG (HSWAG), which is a more challenging benchmark \cite{zellers2018swag}.

\subsubsection{Adversarial Robustness}
We utilized the same models and training procedures as the in-domain evaluation. The settings for adversarial attack follow those from \citet{morris2020textattack} with maximum sentence length $512$.

\subsection{Neural Machine Translation}

\subsubsection{Model Specifications}
Following the Neural Machine Translation (NMT) setting from \citet{deng2018latent}, we utilize the bidirectional LSTM to embed each source sentence to source representations. Attention is utilized, during the decoding stage, to identify which source positions should be used to predict the target using a function of previous generated tokens as the query. The aggregated features are passed to an MLP to produce the distribution over the next target word (see details in \citet{deng2018latent}).

\subsubsection{Experimental Settings}
For NMT we use the IWSLT dataset \cite{cettolo2014report}. We follow the same preprocessing as
in \citet{edunov2017classical} which uses Byte Pair Encoding vocabulary over the combined source/target training set to obtain a vocabulary size of $14$k tokens \cite{sennrich2015neural} with sequences of length up to $125$. 
A two-layer bi-directional LSTM with $512$ units is used as the encoder and another two-layer LSTM with $768$ units is used as the decoder. Other training details include: the batch size $6$, dropout rate $0.3$, and learning rate $3e^{-4}$ with Adam optimizer \cite{kingma2014adam}. During testing, we use beam search with beam size $10$ and length penalty as 1 \cite{wu2016google}.


\subsection{Visual Question Answering}
\subsubsection{Model Specifications}

The state-of-the-art VQA model, MCAN \cite{yu2019deep}, is used in the experiments. The MCAN consists of MCA layers. Each MCA layer consists of self-attention (SA) over question and image features, and guided-attention (GA) between question and image features. Multi-head structure as in \citet{vaswani2017attention}, including the residual and layer normalization components, is incorporated in the MCA layer. MCAN represents the deep co-attention model
which consists of multiple MCA layers cascaded in depth to gradually refine the attended image and question features. We adopt the encoder-decoder structure in MCAN \cite{yu2019deep} with four co-attention layers.

\subsubsection{Experimental Settings} 
We conduct experiments on the commonly used benchmark, VQA-v2 \citep{goyal2017making}, containing human-annotated question-answer (QA) pairs. There are three types of questions: Yes/No, Number, and Other.
The dataset is split into the training (80k images and 444k QA pairs), validation (40k images and 214k QA pairs), and testing (80k images and 448k QA pairs) sets. We perform evaluation on the validation set as the true labels for the test set are not publicly available \citep{deng2018latent}.
To construct the noisy dataset, we incorporate the Gaussian noise (mean 0, variance 5) to image features. We use the same model hyperparameters and training settings in \citet{yu2019deep} as follows:
the dimensionality of input image features, input question features, and fused multi-modal features are set to be $2048$, $512$, and $1024$, respectively. The latent dimensionality in the multi-head attention is $512$, the number of heads is set to $8$, and the latent dimensionality for each head is $64$. The size of the answer vocabulary is set to $N = 3129$ using the strategy in \citet{teney2018tips}. To train the MCAN model, we use the Adam optimizer \citep{kingma2014adam} with $\beta_1 = 0.9$ and $\beta_2 = 0.98$. The base learning rate is set to $\min(2.5te^{-5}, 1e^{-4})$, where $t$ is the current epoch number starting from $1$. After $10$ epochs, the learning rate is decayed by $1/5$ every $2$ epochs. All the models are trained up to $13$ epochs with the same batch size~of~$64$.

\subsubsection{Ablation Study}
\begin{table}[htp!]
\vspace{-4mm} 
\caption{\small Ablation study of the upward path in BABN on VQA.}
\centering
\label{tab:vqa_ablation}
\begin{sc}
\resizebox{1\columnwidth}{!}{
\begin{tabular}{@{}lccccc@{}}\toprule
\small  & \multicolumn{2}{c}{\small Accuracy $\uparrow$} && \multicolumn{2}{c}{\small PAvPU $\uparrow$}\\
\cmidrule{2-3} \cmidrule{5-6}
\small  & \small Original & \small Noisy &&\small Original 
& \small Noisy \\\midrule
\small $\rho=0$, $\sigma$=1.00$e^{-6}$  & 44.62\small & 32.16 && 50.93 & 53.22 \\
\midrule
\small $\rho=1.5$, $\sigma$=0 & 66.78\small & 64.04 && 69.99 & 69.02\\
\midrule
\small $\rho=1.5$, $\sigma$=1.00$e^{-6}$ & {\bf66.92} & {\bf64.40} && {\bf72.21} & {\bf70.43}\\
\bottomrule
\end{tabular}}
\end{sc} 
\end{table}

We conduct ablation study to exam the role of the upward-downward structure by turning the weight parameters $\rho$ and $\sigma$ to zeros. Table~\ref{tab:vqa_ablation} shows that tuning either parameter to zero would lead to performance drop, especially the parameter $\rho$, which demonstrates the necessity and effectiveness of the upward-downward structure. 
We also found that the experimental results are not sensitive to the choice of the value of the $\rho$. Any number from 1 to 4 would give similar results. The other is the scaling factor $\sigma$ that controls the importance of the $\hv^l$ in $\lambdav^l$. We found that the performance is not that sensitive to its value and it is often beneficial to make it smaller. In all experiments considered in the paper, which cover various noise levels and model sizes, we have simply fixed it at 1.00$e^{-6}$.

\end{document}